\documentclass[journal]{IEEEtran}
\usepackage{times}
\usepackage{epsfig}
\usepackage{graphicx}
\usepackage[cmex10]{amsmath}
\usepackage{amssymb}
\usepackage{array}
\usepackage{subfigure}
\usepackage{makecell,rotating,multirow,diagbox}
\usepackage{array}
\usepackage{mathrsfs}
\usepackage{color}
\usepackage[boxed]{algorithm2e}
\usepackage{booktabs}
\newcolumntype{I}{!{\vrule width 1pt}}
\newcolumntype{L}{!{\vrule width 0.5pt}}
\newlength\savedwidth

\newcommand\whline{\noalign{\global\savedwidth\arrayrulewidth
                            \global\arrayrulewidth 1.0pt}%
                  \hline
                   \noalign{\global\arrayrulewidth\savedwidth}}
\newcommand\mhline{\noalign{\global\savedwidth\arrayrulewidth
                            \global\arrayrulewidth 0.5pt}%
                   \hline
                   \noalign{\global\arrayrulewidth\savedwidth}}

\newcommand{\ignore}[1]{}

\usepackage[pagebackref=false,breaklinks=true,letterpaper=true,colorlinks,citecolor=blue,linkcolor=blue,bookmarks=false]{hyperref}
\SetAlCapSkip{0.5em}
\begin{document}

\title{Egocentric Hand Detection Via Dynamic Region Growing}

\author{Shao~Huang,~\IEEEmembership{Member,~IEEE,} \ \
        Weiqiang~Wang,~\IEEEmembership{Member,~IEEE,} \ \
        Shengfeng~He,~\IEEEmembership{Member,~IEEE,} \ \
        and \ \ Rynson~W.H.~Lau,~\IEEEmembership{Senior Member,~IEEE}
\thanks{Shao Huang is with the University of Chinese Academy of Sciences and City University of Hong Kong. E-mail: shaohuang6-c@my.cityu.edu.hk.}
\thanks{Weiqiang Wang is with the University of Chinese Academy of Sciences, China. Email: wqwang@ucas.ac.cn.}
\thanks{Shengfeng He is with the School of Computer Science and Engineering, South China University of Technology, China. E-mail: hesfe@scut.edu.cn.}
\thanks{Rynson W.H. Lau is with the Department of Computer Science, City University of Hong Kong, Hong Kong. E-mail: Rynson.Lau@cityu.edu.hk.}
}

\maketitle

\begin{abstract}
Egocentric videos, which mainly record the activities carried out by the users of the wearable cameras, have drawn much research attentions in recent years. Due to its lengthy content, a large number of ego-related applications have been developed to abstract the captured videos. As the users are accustomed to interacting with the target objects using their own hands while their hands usually appear within their visual fields during the interaction, an egocentric hand detection step is involved in tasks like gesture recognition, action recognition and social interaction understanding. In this work, we propose a dynamic region growing approach for hand region detection in egocentric videos, by jointly considering hand-related motion and egocentric cues. We first determine seed regions that most likely belong to the hand, by analyzing the motion patterns across successive frames. The hand regions can then be located by extending from the seed regions, according to the scores computed for the adjacent superpixels. These scores are derived from four egocentric cues: contrast, location, position consistency and appearance continuity. We discuss how to apply the proposed method in real-life scenarios, where multiple hands irregularly appear and disappear from the videos. Experimental results on public datasets show that the proposed method achieves superior performance compared with the state-of-the-art methods, especially in complicated scenarios.
\end{abstract}


 \ifCLASSOPTIONpeerreview
 \begin{center} \bfseries EDICS Category: 3-BBND \end{center}
 \fi
%
\IEEEpeerreviewmaketitle

\section{Introduction}

In recent years, wearable cameras like {the} Google Glass are popular for capturing egocentric videos and have been used to record the daily life {of the camera wearer} in a first person point of view. The popularity of egocentric videos leads to novel applications {such as} gaze prediction~\cite{li2013learning}, gaze analysis in multi-party conversations~\cite{kumano2016}, summarization~\cite{ghosh2012discovering,lu2013story}, finger tracking~\cite{dominguez2006robust}, snap point prediction~\cite{xiong2014detecting}, facial attribute representation~\cite{Wang_2016_CVPR}, {and photographer identification~\cite{Hoshen_2016_CVPR}}. {The hands play a major role in a lot of applications, as humans are accustomed to interacting with objects using the} hands~\cite{fathi2011learning}. This makes egocentric hand detection an essential task for egocentric video analysis.

\begin{figure}[t]
\centering
\subfigure{
\includegraphics[width=0.225\textwidth]{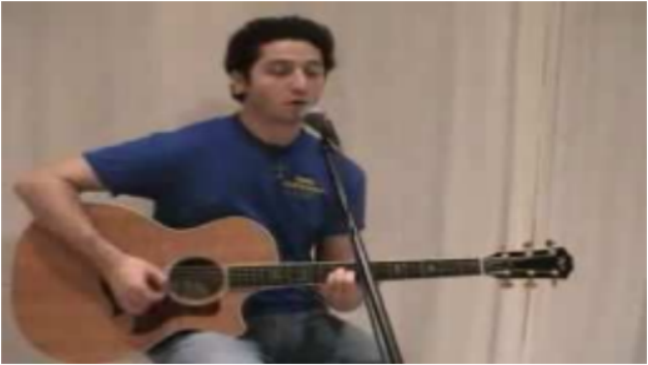}
}
\subfigure{
\includegraphics[width=0.225\textwidth]{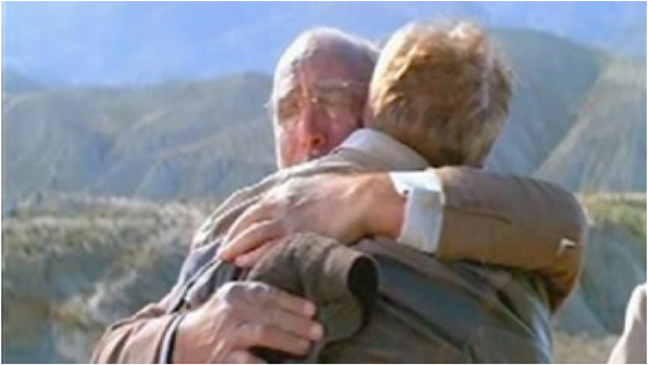}
}
\subfigure{
\includegraphics[width=0.225\textwidth]{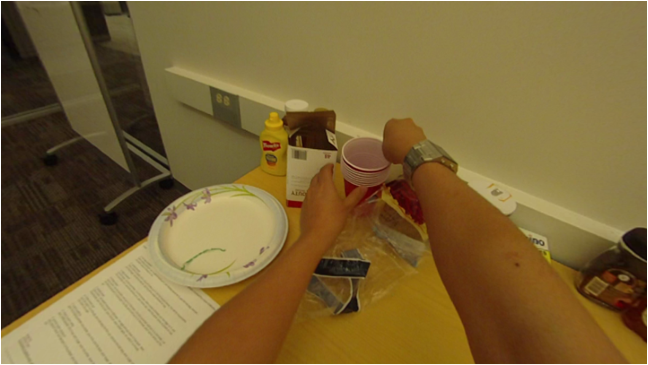}
}
\subfigure{
\includegraphics[width=0.225\textwidth]{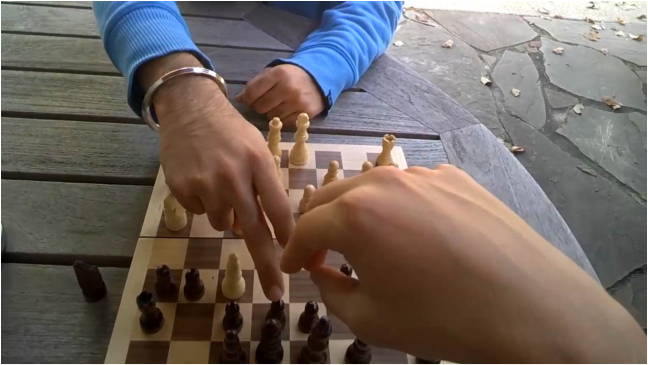}
}
\caption{Hand detection in hand-held and egocentric videos. Left: {two} frames from hand-held videos. Right: {two} frames from egocentric videos.}
\label{fig:general_egocentric}
\end{figure}

To accurately locate the appeared hands, various methods have been proposed, either based on hand-crafted cues~\cite{fathi2011learning,li2013model} or deep architecture~\cite{bambach2015lending}. These works are designed for different scenarios, e.g., {indoor~\cite{fathi2011learning,li2013model}, outdoor~\cite{li2013pixel,ren2010figure} scenes, or social interactions involving multiple people}~\cite{bambach2015lending,lee2014hand}. {In general, algorithms that can handle complex environments involving multiple persons impose high computational costs~\cite{bambach2015lending} in order to achieve satisfactory results. Algorithms that can only handle simple scenes likely fail in real-life scenarios~\cite{ren2010figure}. In other words}, existing methods cannot achieve a good balance between efficiency and accuracy. This problem motivates us to {develop a new method that produces satisfactory performance at a low computational cost, especially in complex} environments.

In this work, {instead of assuming that only the wearer's hands appearing in the scene~\cite{li2013pixel,ren2010figure}, we observe that camera and hand motion patterns have distinct properties, which can be used to separate the hands from the video content. This is achieved by constructing an appropriate homography matrix among the correspondences of the} successive frames. The seed regions, which {are parts of the hands, are identified} based on these motion patterns. The hand regions {can then be progressively determined by dynamic region growing, starting from the seeds. The region growing algorithm is} based on the proposed egocentric cues: contrast, location, position consistency and appearance continuity. {We also discuss implementation issues on how to apply this work in complicated environments.}

The main contributions of the proposed approach are summarized as follows:
\begin{itemize}
  \item {We propose an algorithm to automatically generate seed regions, by differentiating camera and hand motion patterns across successive frames. As these seeds are adaptively identified from the video itself, there is no need to make  assumptions such as} static background or limited number of hands.
  \item {We propose a region growing algorithm by adding one adjacent superpixel} with the highest score in each iteration. {This} score is computed based on the appearance and spatial constraints, in the {current and} previous frames.
  \item {We propose to dynamically initialize, update and expire the appearance models to improve the hand detection in complex environments at low computational cost}.
\end{itemize}

The rest of this paper is organized as follows. Section~\ref{sec:related} introduces related works. Section~\ref{sec:method} presents the proposed approach for egocentric hand detection {and discusses implementation issues}. Section~\ref{sec:experiments} {evaluates the proposed method on public datasets. Finally, Section~\ref{sec:conclusion} draws a conclusion of this work.}

\section{Related Work}\label{sec:related}
In this section, we first briefly summarize the latest progress made on egocentric video analysis. {We then discuss} related works on egocentric hand detection.

\subsection{Egocentric Video Analysis}

{Egocentric videos are captured by a head-mounted camera. {Thus, they largely differ from hand-held videos in many aspects. As the camera wearer moves his head, the content of the captured video changes,} which indirectly reflects the wearer's actions. This distinct feature recently leads to many interesting research problems on egocentric videos. {Some survey papers, e.g., \cite{betancourt2015evolution} and Bambach~\cite{bambach2015survey}, are published to summarize the latest progress. They explain how egocentric video analysis deals with the classical problems and its novel challenges, by dividing these works into multiple categories according to the various} objectives. Recently, Molino \emph{et al.}~\cite{del2017summarization} compare {the latest segmentation methods and selection algorithms. The popular egocentric datasets and evaluation metrics are also discussed to provide a comprehensive review. All the existing} methods aim to analyze and understand egocentric videos by abstracting the content in different aspects (from the perspective of the wearer), thus {saving} human efforts in {the time-wasting process of} browsing the entire videos. {Our} work shares a similar objective via improving and accelerating egocentric hand detection, to benefit gesture recognition~\cite{baraldi2014gesture}, pose recognition~\cite{rogez2015first}, action recognition~\cite{li2015delving} and understanding social interactions~\cite{bambach2015lending}.}

\begin{figure*}[!htbp]
\centering
\subfigure[All correspondences $\mathbf{M}$]{
\includegraphics[width=0.4\textwidth]{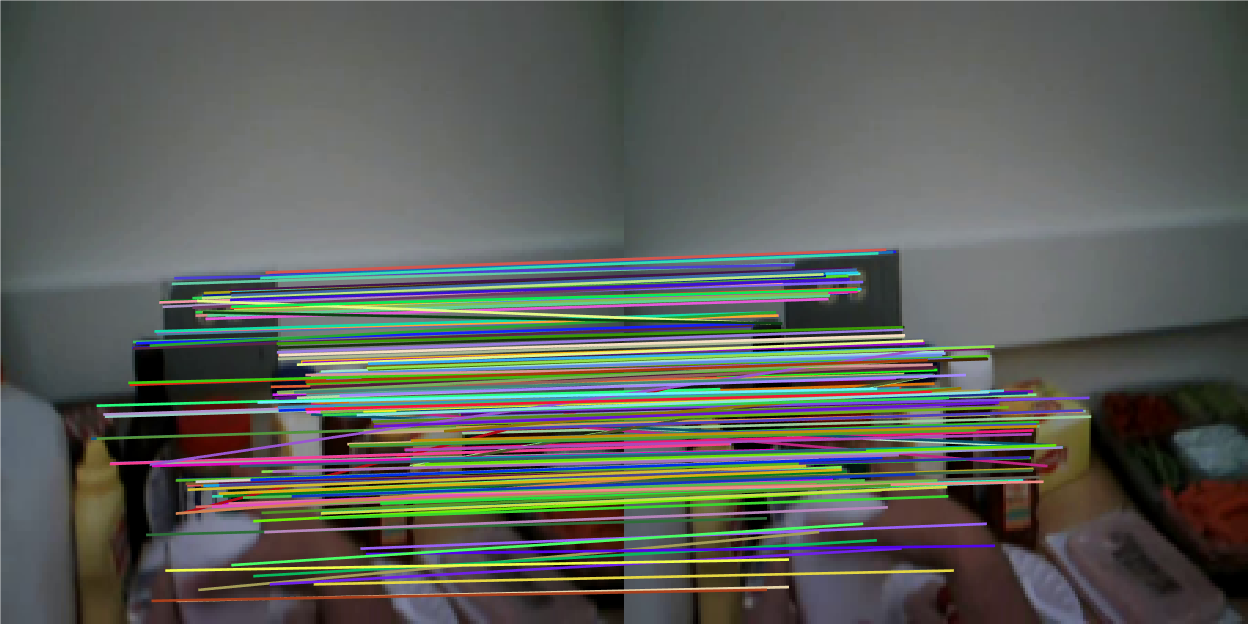}
}
\hspace{3mm}
\subfigure[Candidate correspondences $\mathbf{M}_r$]{
\includegraphics[width=0.4\textwidth]{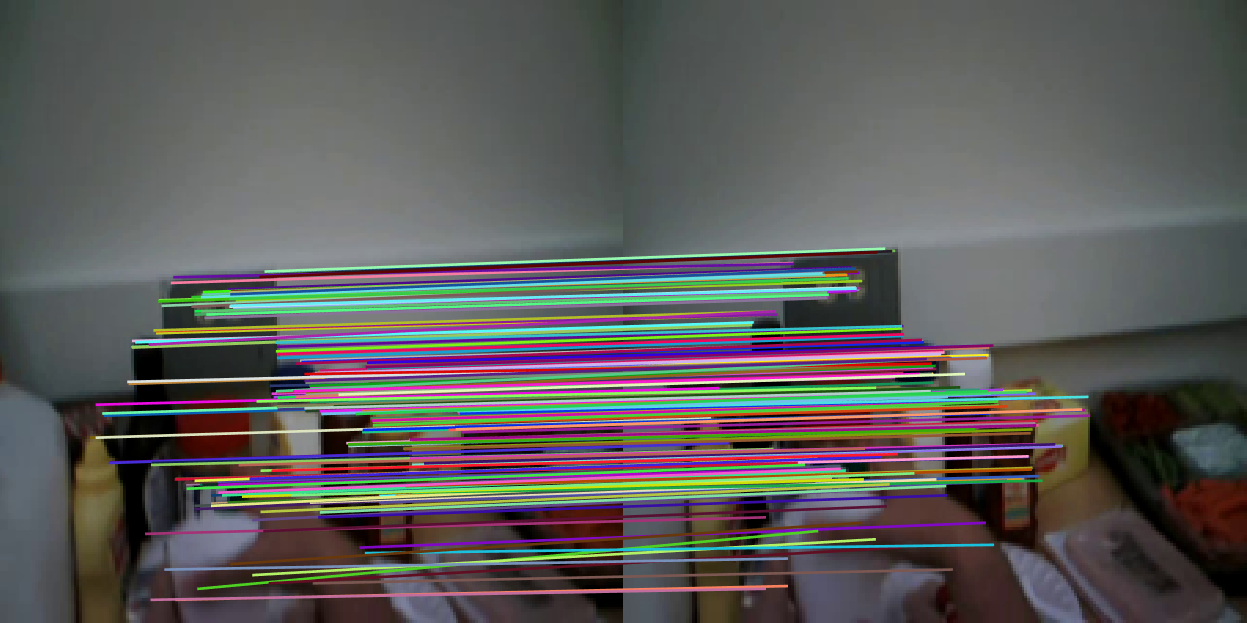}
}
\\
\subfigure[Hand-related correspondences $\mathbf{M}_h$]{
\includegraphics[width=0.4\textwidth]{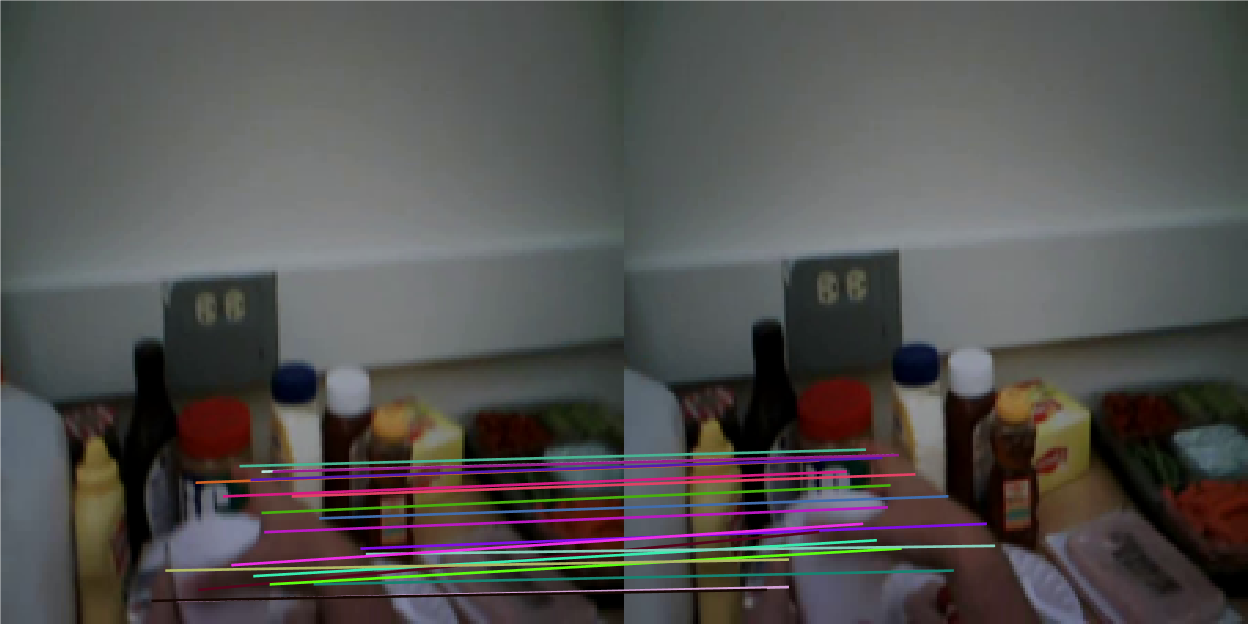}
}
\caption{Steps for {computing} hand-{related} correspondences. The solid lines with different colors denote the correspondences {across two} successive frames.}
\label{fig:correspondences}
\end{figure*}

\subsection{Egocentric Hand Detection}
As shown in Fig.~\ref{fig:general_egocentric}, {the hands appear differently in hand-held and egocentric videos. Many researchers have recently made great efforts to investigate egocentric hand detection, by suggesting interesting assumptions and addressing the problem in confined environments, such as a single person performing daily activities indoors~\cite{fathi2011learning,li2013model} or outdoors~\cite{li2013pixel,ren2010figure}, or multiple persons involved in social interactions~\cite{bambach2015lending,lee2014hand}. Ren and Gu~\cite{ren2010figure} are among the first to focus on hand detection in egocentric videos. They regard the problem as figure-ground segmentation, and locate hand regions by assuming that the hands have irregular optical flow patterns compared with the} background. Fathi \emph{et al.}~\cite{fathi2011learning} present the Georgia Tech Egocentric Activity (GTEA) dataset and conduct hand detection by incorporating multiple color features to segment {the} hands from objects. Li and Kitani~\cite{li2013model} regard hand detection as a model recommendation task, where the n-best hand detectors are recommended based on the probe set, referring to a small amount of labeled data from the test distribution. To further address the limitation brought by the probe set, the virtual probes are proposed which can be automatically extracted from the test distribution. Another work {presented by Li and Kitani~\cite{li2013pixel} proposes} a fully labeled indoor/outdoor egocentric hand detection benchmark dataset containing over 200 million labeled pixels, and gives extensive analysis of detection performance using a wide range of local appearance features. Lee \emph{et al.}~\cite{lee2014hand} model {all hands in a social interaction instead of just} the wearer's hands, and encode spatial arrangements to disambiguate hand types using a probabilistic graphical model. Bambach \emph{et al.}~\cite{bambach2015lending} further investigate the social interactions on their dataset named EgoHands, {which} contains 48 first-person videos of people interacting in realistic environments and also provides pixel-level ground truth for over 15000 hand instances. They use strong appearance models with Convolutional Neural Networks (CNNs), and introduce a simple candidate region generation approach to achieve state-of-the-art performance. Among the above works, \cite{fathi2011learning,ren2010figure} assume that the background is static so that the optical flow patterns {can be used} for segmentation. Since the assumption is not true in real-world egocentric videos where {the wearers likely walk around, it is difficult to achieve a} satisfactory performance. Besides, \cite{li2013pixel,li2013model} assume no social interactions involved. {Thus,} their system generally performs poorly in scenes where the wearer is interacting with the partner, e.g., playing cards in the room. Although the deep-based method~\cite{bambach2015lending} achieves promising performance in {complex} environments, the disadvantage is the high computational cost and {it requires a large amount of data} for training. Hence, the performance is greatly degraded when insufficient training data is available in specific environments.

{In contrast, our work does not assume a confined environment or the persons involved. The proposed method learns this information directly from the video. Our observation is that the hands are frequently moving and interacting with objects. As their motions are different from head motions, our method can obtain the seed regions with high confidence. By jointly considering appearance and spatial constraints, it can progressively locate the hand regions.}

\section{Methodology}\label{sec:method}
{Our region growing approach for egocentric hand detection is based on gradually extending the seed regions into hand regions. To improve computation efficiency, it is built on superpixels, instead of pixels~\cite{li2013pixel}, as motivated by our} observation that the over-segmentation method like~\cite{achanta2012slic} could accurately detect the boundaries between hand regions and the background. For the rest of this section, {we first identify the seed regions by differentiating camera and hand motion patterns across successive frames in Section~\ref{sec:method_seed}. We then present the iterative procedure for extending the seed regions into hand regions in Section~\ref{sec:method_growth}. Finally, we discuss some} implementation {issues} in Section~\ref{sec:method_discussion}.

\subsection{Seed Region Generation}\label{sec:method_seed}

Egocentric videos differ significantly from hand-held videos. The movement and the {change of viewpoint are more unpredictable, since the motion of the wearer's head, like swinging, shaking and nodding, may occur anytime}. {Besides, while moving objects in the background may frequently appear in hand-held videos~\cite{kolsch2004fast} and some egocentric videos captured in {outdoor scenes}, the current egocentric datasets {for} detection/recognition focus on {what the wearers are doing} and are captured mostly indoors, where moving background objects are indeed rare~\cite{pirsiavash2012detecting,fathi2012learning,fathi2011learning,li2013pixel}.} As suggested by~\cite{fathi2011learning,li2013pixel}, the {motions in egocentric videos are mainly produced} by the mounted camera and the hands. Hence, {we first establish the candidate correspondences across successive frames, and then classify the hand-related correspondences, which} distribution leads to seed regions. Fig.~\ref{fig:correspondences} shows the steps for {determining the hand-related} correspondences.

In the preprocessing step, a given frame is first over-segmented using SLIC~\cite{achanta2012slic}, which is known as simple linear iterative clustering, and adapts a K-means clustering approach to efficiently generate superpixels.  {The ORB~\cite{rublee2011orb} descriptors are then computed across successive frames. These descriptors are rotation invariant and resistant to noise, and can produce} real-time performance. The extraction of {these descriptors} is much faster compared with SIFT ~\cite{lowe2004distinctive} and SURF ~\cite{bay2006surf} descriptors. For a pair of successive frames $p$ and $q$, the generated correspondences form the set $\mathbf{M}=\{\mathbf{m}_i|\mathbf{m}_i=(\mathbf{p}_i,\mathbf{q}_i),i=1,2,\cdots,N \}$, where $\mathbf{p}_i$ and $\mathbf{q}_i$ denote the $i$th 2D matching coordinates in $p$ and $q$, respectively, as shown in Fig.~\ref{fig:correspondences}(a).

{Despite the efficiency of the ORB descriptors, some false correspondences may be introduced, which could be {neglected} because of the subtle motion caused by the small time interval, e.g., 1/30 of a second for the EgoHands dataset~\cite{bambach2015lending}. Thus, the set of obviously false correspondences $\mathbf{M}_f$ is considered as noise and directly excluded. Therefore,}
\begin{eqnarray}
    \mathbf{m}_k \in \mathbf{M}_f, \ s.t. \ \|\mathbf{p}_k-\mathbf{q}_k\|_2 > \theta, \ k=1,2,\cdots,N,
\end{eqnarray}
where $\theta$ is the largest position difference allowed. {It is adaptively learned for each video sequence, instead of for a single frame, as the behavior of the camera wearer, the environment and the configuration of the wearable camera tend to be consistent over the entire video}. Suppose {that} the median value of all position differences in the video is $\nu$. {Then,} $\theta=10 \nu$ is set in our system to differentiate false correspondences. Thus, the set of candidate correspondences $\mathbf{M}_r$ is generated. {Therefore,}
\begin{eqnarray}
    \mathbf{M}_r=\mathbf{M} \backslash \mathbf{M}_f.
\end{eqnarray}
It precisely reflects the motion between $p$ and $q$, as shown in Fig.~\ref{fig:correspondences}(b). For {the} camera motion pattern, all objects in the background are affected, e.g., bottles, chess, plates, as the visual field is determined by the head-mounted camera. {For the hand motion pattern,} only hand regions are affected. Hence, more correspondences agree with {the} camera motion pattern and distribute in the whole frame, while {the hand-related} correspondences distribute in a few regions. We calculate the homography matrix from the candidate correspondences with RANSAC~\cite{fischler1981random}, {and consider the camera-related correspondences $\mathbf{M}_c$ as} "inliers".

Hence, the set of hand-relevant correspondences $\mathbf{M}_h$ is formulated as
\begin{eqnarray}
    \mathbf{M}_h=\mathbf{M}_r \backslash \mathbf{M}_c.
\end{eqnarray}
These correspondences denote where {the} hands possibly lie, as shown in Fig.~\ref{fig:correspondences}(c). Unfortunately, this method may fail to extract all hand-{related} correspondences, and may occasionally contain {noise} when the camera and hand motion patterns are indistinguishable in some frames. Therefore, it is {not suitable} to directly regard the superpixels containing $\mathbf{m} \in \mathbf{M}_h$ as hand regions. Instead, we locate {the} seed regions based on the distribution of $\mathbf{M}_h$.

\begin{figure}[!htbp]
\centering
\subfigure[Seed region (green)]{
\includegraphics[width=0.22\textwidth]{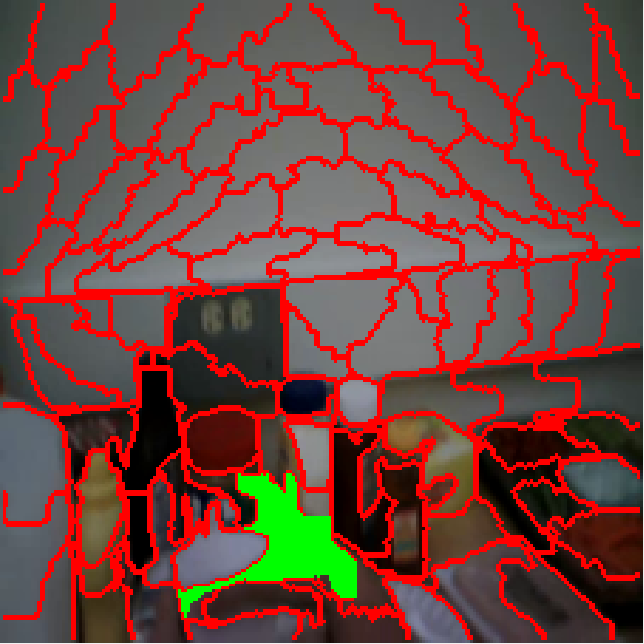}
}
\hspace{-3mm}
\subfigure[Located hand regions (blue)]{
\includegraphics[width=0.22\textwidth]{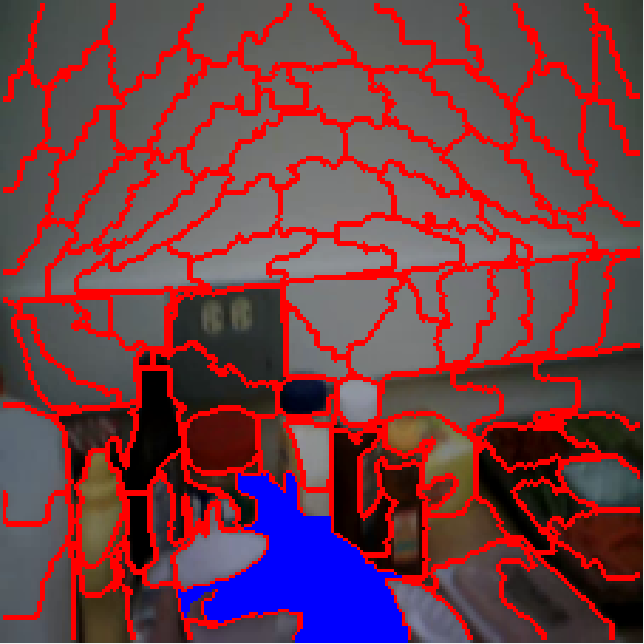}
}
\caption{The calculated seed region and the located hand regions.}
\label{fig:hand}
\end{figure}

Suppose a set of superpixels is obtained, i.e., $\mathbf{S}=\{\mathbf{s}_k| k=1,2,\cdots,L\}$ after the {over-segmentation step} using SLIC~\cite{achanta2012slic} for each frame, and superpixel $\mathbf{s}_k$ contains $\lambda_k$ hand-{related} correspondences, $ \ k=1,2,\cdots,L$. The neighbors of a superpixel are defined as adjacent superpixels sharing parts of {the} boundaries. $\mathbf{s}_k$ is regarded as a peak if $\lambda_k$ is the largest among all its neighbors and {the} neighbors' neighbors, to avoid multiple peaks in a local region. For these peaks, $\lambda_k \geq 0.1 \mu$ is set to remove {noise, which contains only few hand-related} correspondences, and $\mu$ is adaptively assigned the average number of hand-{related} correspondences of all frames in {the input} video sequence. The retained peaks form the set of seed regions $\mathbf{\hat{S}}=\{\mathbf{\hat{s}}_u| u=1,2,\cdots,T\}$, as shown in Fig.~\ref{fig:hand}(a). Each seed $\mathbf{\hat{s}}_u \in \mathbf{\hat{S}}$ is assumed to have the priority to {form a part of the hand}.

\subsection{Hand Region Growing}\label{sec:method_growth}
{To locate the hand regions, we further utilize color and position information by jointly considering four egocentric cues: contrast, location, position consistency and appearance continuity.} These cues are proposed based on appearance and spatial constraints, in the current frame as well as previous frames:

\begin{itemize}
\item \textbf{Contrast}: The contrast cue aims to build the appearance constraint in the current frame, between seeds $\mathbf{\hat{S}}$ and superpixels $\mathbf{S}$. A given frame is first converted into color space HSV. {The normalized color histogram $\mathbf{h}=[h_1,h_2,\cdots,h_P]$ with $P$ bins is then} calculated for each superpixel, where $P=16$ in our system for the sake of low computational cost. {We use 8 bins for the H channel and 4 bins for S and V channels. These bins are stacked to form the histogram.} The score $S_{1}(\mathbf{s}_k, \mathbf{\hat{s}}_u)$ for superpixel $\mathbf{s}_k \in \mathbf{S}$ and seed $\mathbf{\hat{s}}_u \in \mathbf{\hat{S}}$ is calculated based on the relative entropy as:
    \begin{eqnarray}\label{equ:contrast}
      S_{1}(\mathbf{s}_k, \mathbf{\hat{s}}_u) = \exp(- \sum_{j=1}^{P} (& h_{k,j} \log\frac{h_{k,j}}{\hat{h}_{u,j}} + \notag \\ 
      & \hat{h}_{u,j} \log\frac{\hat{h}_{u,j}}{h_{k,j}})),
    \end{eqnarray}
    where $h_{k,j}$ and $\hat{h}_{u,j}$ {denote the $j$th bin of the histogram on $\mathbf{s}_k$ and $\mathbf{\hat{s}}_u$,} respectively.

\item \textbf{Location}: The location cue aims to build the spatial constraint in the current frame, between seed $\mathbf{\hat{s}}_u \in \mathbf{\hat{S}}$ and superpixel $\mathbf{s}_k \in \mathbf{S}$. Suppose the centers of $\mathbf{\hat{s}}_u$ and $\mathbf{s}_k$ are denoted by $\mathbf{\hat{c}}_u$ and $\mathbf{c}_k$, respectively. {This score $S_{2}(\mathbf{s}_k, \mathbf{\hat{s}}_u)$ is then defined as:}
    \begin{eqnarray}\label{equ:location}
      S_{2}(\mathbf{s}_k, \mathbf{\hat{s}}_u) = \exp(- \frac{\|\mathbf{c}_k - \mathbf{\hat{c}}_u\|_2}{l}),
    \end{eqnarray}
    where $l$ is the length of frame diagonal, and $\|\cdot\|_2$ denotes the L2 norm;

\item \textbf{Position Consistency}: As egocentric hand detection is conducted through the whole video instead of a single frame, temporal information {helps} build the spatial constraint {between} successive frames, by taking position consistency into account. For superpixel $\mathbf{s}_k \in \mathbf{S}$ and any $\mathbf{\bar{s}}'_{w} \in \mathbf{\bar{S}}'$ which denotes the calculated hand regions in the previous frame, the score $S_{3}(\mathbf{s}_k, \mathbf{\bar{s}}'_{w})$ is defined as:
    \begin{eqnarray}\label{equ:position}
      S_{3}(\mathbf{s}_k, \mathbf{\bar{s}}'_{w}) = \exp(\frac{\|\mathbf{c}_k - \mathbf{\bar{c}}'_w\|_2}{l}),
    \end{eqnarray}
    where $\mathbf{\bar{c}}'_w$ denotes the center of $\mathbf{\bar{s}}'_{w}$.

\item \textbf{Appearance Continuity}: For egocentric videos, we observe that the {appearance of a specific hand has only minor change} over time, because the camera wearer would not be replaced and the illumination tends to be stable in a short period of time, as suggested by~\cite{pirsiavash2012detecting,fathi2012learning,fathi2011learning,bambach2015lending,li2013pixel}. Multiple appearance models are built and updated as multiple hands may irregularly appear in the video. Similar to the contrast cue, the normalized color histogram is built for modeling appearance continuity. For superpixel $\mathbf{s}_k \in \mathbf{S}$ and any $\mathbf{a}_v^{t-1} \in \mathbf{A}^{t-1}$ which denotes the accumulated model at previous timestamp $t-1$, the score $S_{4}(\mathbf{s}_k,\mathbf{a}_v^{t-1})$ is defined as:
    \begin{eqnarray}\label{equ:appearance}
      S_{4}(\mathbf{s}_k, \mathbf{a}_v^{t-1}) = \exp(- \sum_{j=1}^{P} (& h_{k,j} \log\frac{h_{k,j}}{a^{t-1}_{v,j}} + \notag \\ 
      & a^{t-1}_{v,j} \log\frac{a^{t-1}_{v,j}}{h_{k,j}})),
    \end{eqnarray}
    where $a^{t-1}_{v,j}$ denotes the $j$th bin of the $v$th appearance model till previous timestamp $t-1$ , and {will be} discussed in Section~\ref{sec:method_discussion}.
\end{itemize}

With the formulations for these proposed cues, score fusion is conducted to measure {the likelihood of $\mathbf{s}_k$ being part of the hand}. {The contrast and location cues are considered jointly to avoid inconsistency.} {Hence, we jointly maximize $S_{1}(\mathbf{s}_k, \mathbf{\hat{s}}_u)$ and $S_{2}(\mathbf{s}_k, \mathbf{\hat{s}}_u)$} to select the optimal seed for $\mathbf{s}_k$. $S_{3}(\mathbf{s}_k, \mathbf{\bar{s}}'_{w})$ and $S_{4}(\mathbf{s}_k,\mathbf{a}_v^{t-1})$ are maximized separately as $\mathbf{\bar{s}}'_{w}$ is independent of $\mathbf{a}_v^{t-1}$. {Unlike} the method proposed by Selective Search~\cite{uijlings2013selective}, where binary weights are used for score fusion, we adopt the weighted summation as:

\begin{eqnarray}\label{equ:fusion}
  S(\mathbf{s}_k) &=  \max \limits_{\mathbf{\hat{s}}_u \in \mathbf{\hat{S}}} [\kappa_1 S_{1}(\mathbf{s}_k, \mathbf{\hat{s}}_u) + \kappa_2 S_{2}(\mathbf{s}_k, \mathbf{\hat{s}}_u)] +\notag \\
                    & \max \limits_{\mathbf{\bar{s}}'_{w} \in \mathbf{\bar{S}}'} \kappa_3 S_{3}(\mathbf{s}_k, \mathbf{\bar{s}}'_{w})
                    + \max \limits_{\mathbf{a}_v^{t-1} \in \mathbf{A}^{t-1}} \kappa_4 S_{4}(\mathbf{s}_k,\mathbf{a}_v^{t-1}),
\end{eqnarray}

where $\kappa_1=\kappa_4=0.3$ and $\kappa_2=\kappa_3=0.2$ are empirically set.

{The hand} regions are gradually located by extending from the seed regions $\mathbf{\hat{s}}_u \in \mathbf{\hat{S}}$. We add the one with the highest score among the remaining adjacent superpixels in each iteration. Algorithm~\ref{alg:at_growth} {summarizes the proposed method. The} impacts of parameters $\alpha$ and $\beta$ are analyzed in Section~\ref{sec:experiments_parameter}. Fig.~\ref{fig:hand}(b) shows an example of {the hand regions obtained}.

\begin{algorithm}[tb]
\caption{Pseudo-code for Hand Region Growing.}\label{alg:at_growth}
\KwIn{each seed $\mathbf{\hat{s}}_u \in \mathbf{\hat{S}}$ with the corresponding score $S(\mathbf{\hat{s}}_u)$;}
\KwOut{the set of hand regions $\mathbf{\bar{S}}$;}

initialize $\mathbf{\bar{S}} = \mathbf{\hat{S}}$, $\eta=\max \limits_{u=1,\cdots,T} S(\mathbf{\hat{s}}_u)$, $i=0$;

\While {true}
{
    update the set of adjacent superpixels $\mathbf{N(\bar{S})}$ of $\mathbf{\bar{S}}$;

    calculate the largest score $\epsilon = \max \limits_{\mathbf{s} \in \mathbf{N(\bar{S})}} S(\mathbf{\mathbf{s}})$;

    \If {$\epsilon < \alpha \times \eta$}
    {
        break;
    }
    \Else
    {
        update $i \leftarrow i+1$;

        select the superpixel $\mathbf{s}_{0}=\arg\max \limits_{\mathbf{s} \in \mathbf{N(\bar{S})}} S(\mathbf{\mathbf{s}})$;

        update $\mathbf{\bar{S}} \leftarrow \mathbf{\bar{S}} \cup \{\mathbf{s}_{0}\}$;

        update $\eta=\epsilon$;
    }
}

{calculate connected components among $\mathbf{\bar{S}}$;}

{remove tiny ones of less than $\beta$ pixels;}
\end{algorithm}

\subsection{{Implementation Issues}}\label{sec:method_discussion}
{Here, we discuss four implementation issues of the proposed method:}

\begin{itemize}
\item \textbf{Appearance Update}: We formulate {the} appearance update based on the {assumption that the contribution of a frame is related to its distance from the current timestamp in the} temporal domain. As suggested by~\cite{bambach2015lending,lee2014hand}, {if multiple hands belonging to different persons appear in a social interaction, they typically have different appearances. Thus, connected components obtained by Algorithm~\ref{alg:at_growth} contribute only to the} relevant models. Concretely, each component in {the} current frame at timestamp $t$ is represented by normalized histogram $\mathbf{\bar{h}}^t$ with $P$ bins, and is assigned to the most similar model based on KL divergence. Suppose {that the $v$th model $\mathbf{a}_v^{t-1} \in \mathbf{A}^{t-1}$ has $n_v$ relevant components. It is then updated as:}
    \begin{eqnarray}\label{equ:discussion_appearance}
      \mathbf{a}_v^{t} &=& \delta \frac{1}{n_v} \sum_{i=1}^{n_v} \mathbf{\bar{h}}^t_{v,i} + (1-\delta) \mathbf{a}_v^{t-1},
    \end{eqnarray}
    where $\delta=0.4$ {controls} the decay over time, and $\mathbf{\bar{h}}^t_{v,i}$ is the $i$th relevant component at timestamp $t$ assigned to $\mathbf{a}_v^{t-1}$. For the models {with no assigned relevant components}, $\mathbf{a}_v^{t} = \mathbf{a}_v^{t-1}$ is assigned. For the frame recognized as ``no hand'' {due to having no seeds}, $\mathbf{a}_v^{t} = \mathbf{a}_v^{t-1}$ is directly assigned for all models; {no update nor} region growing is conducted for computational efficiency.

\item \textbf{Dynamic Initialization}: We have no access to the priori at the beginning of a video sequence, {such as hand location bias (i.e., the hands have a higher chance to appear in the central region than the boundary region of a frame), velocity distribution (i.e., the maximum/avarage hand and head motions), and restricted to only having the wearers' hands.} Besides, it is unreasonable to assume that {the} hands would appear in the first frame {or within the visual field all the time}. In {practice, multiple hands may appear and disappear irregularly, such as in~\cite{fathi2011learning}. Thus, the appearance models are dynamically updated and expired to keep only the active models as}:

\begin{itemize}
\item The position consistency term in Eq.~\ref{equ:fusion} is {set to 0 if the previous frame is recognized as having ``no hand''. If seeds are found in the current frame, the hand position will then be initialized by Algorithm~\ref{alg:at_growth}.}

 \item The representation $\mathbf{\bar{h}}^t$ for a connected component at timestamp $t$ may differ largely from any existing model $\mathbf{a}_v^{t-1} \in \mathbf{A}^{t-1}$, e.g., a hand appears in the visual field for the first time. In this case, a new model is initialized to $\mathbf{\bar{h}}^t$, under the condition that $\mathbf{\bar{h}}^t$ is continuously observed with only ``subtle changes'' at successive timestamps $t+1,\cdots,t+10$. Concretely, suppose {that} the average KL divergence of each component from its most similar model at timestamp $t$ is $\sigma^t$, and the KL divergence of $\mathbf{\bar{h}}^t$ from its most similar model is $\rho^t$. {The condition $\rho^t / \sigma^t \geq 3$ is then used to determine whether $\mathbf{\bar{h}}^t$ is distinctive enough to form} a new model, and $\parallel \mathbf{\bar{h}}^t - \mathbf{\bar{h}}^{t+\gamma} \parallel_2 \leq 0.2, \ \gamma=1,\cdots,10$, is defined as the consistent constraint to depict ``subtle changes''. {Note that occasionally {a background region may be falsely recognized as a hand region in a one frame. This, however, would not result in a new appearance model $\mathbf{\bar{h}}^t$, as the error would not propagate to successive frames. Thus, the above conditions would not be} satisfied.}

\item A model is considered to be expired if no {related components are assigned to it for} a period of time, e.g., $500$ successive frames in our system. The expired model is removed from $\mathbf{A}^{1:t}$ {(as a model expiry mechanism) to ensure that all current} models are active.
\end{itemize}

\item \textbf{Egocentric Hand Patterns}: As suggested by~\cite{li2013learning}, four hand patterns may appear in a single egocentric frame. For ``left hand only'', ``right hand only'' and ``intersecting hands'', {all hand regions are normally connected. Hence, fewer seeds would be generated, resulting} in a single connected component using Algorithm~\ref{alg:at_growth}. For ``two separate hands'', the generated seeds would be located in {two different hands, resulting in} two connected components. In this work, {we consider an additional pattern called ``no hand'', as hands may disappear in the visual field, e.g., when the wearer is just sitting in front of a table. In this case, no seeds would be generated, as the condition $\lambda_k \geq 0.1 \mu$ proposed in Section~\ref{sec:method_seed} is not satisfied due to the lack of hand-related} correspondences. Hence, {this frame can be quickly recognized as ``no hand'', without running the region growing algorithm. In addition, we also consider ``multiple hands'', which is introduced by~\cite{bambach2015lending} for} social interactions, e.g., the wearer plays cards with {friends. This hand pattern would produce more seeds distributed in multiple locations. Occasionally, false seeds that do not belong to any hands may be generated, if the camera and hand motion patterns are indistinguishable caused by the wearer's clothes or bystanders. Refinement is then} applied at the end of Algorithm~\ref{alg:at_growth}. {Fig.~\ref{fig:patterns} shows examples of our egocentric hand patterns.}
\begin{figure}[htbp]
\centering
\subfigure[Left hand only]{
\includegraphics[width=0.15\textwidth]{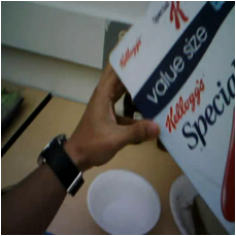}
}
\hspace{-3mm}
\subfigure[Right hand only]{
\includegraphics[width=0.15\textwidth]{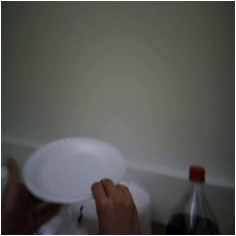}
}
\hspace{-3mm}
\subfigure[Intersecting hands]{
\includegraphics[width=0.15\textwidth]{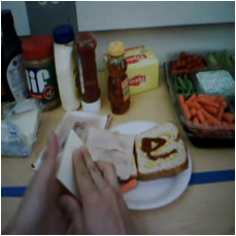}
}
\\
\subfigure[Two separate hands]{
\includegraphics[width=0.15\textwidth]{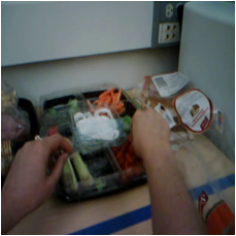}
}
\hspace{-3mm}
\subfigure[No hand]{
\includegraphics[width=0.15\textwidth]{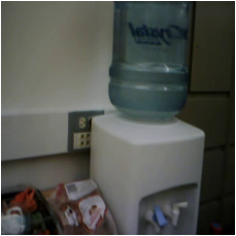}
}
\hspace{-3mm}
\subfigure[Multiple hands]{
\includegraphics[width=0.15\textwidth]{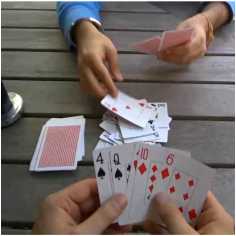}
}
\caption{Egocentric hand patterns.}
\label{fig:patterns}
\end{figure}

\item \textbf{Efficient Computation}: To improve the time efficiency, we benefit from the following techniques: ORB descriptors~\cite{rublee2011orb} for accelerating {the correspondence computation across} successive frames, {computationally efficient histogram representation for modeling appearances, score computation of only the set of adjacent superpixels}, model expiry mechanism for restricting the number of appearance models. Besides, {we also employ an early rejection approach}, following the idea of cascade of classifiers for detection~\cite{cevikalp2013face,betancourt2014sequential}:
    {the condition, $\max_{\mathbf{\hat{s}}_u \in \mathbf{\hat{S}}} S_{1}(\mathbf{s}_k, \mathbf{\hat{s}}_u) \geq 0.5$, is set to efficiently remove the {superpixels that obvious belong to the background superpixels as their appearances differ} largely from the seeds. In indoor scenarios such as {those in} the GTEA~\cite{fathi2011learning} and ADL~\cite{pirsiavash2012detecting} datasets, the illumination does not change significantly. Thus, the contrast cue can be reliably used to exclude some obvious background superpixels which appearances differ largely from the seeds. For those regions that cannot be easily recognized as background or consist of both noise and hand regions, other cues ({i.e.,} location, position consistency and appearance continuity) are further computed for classification.}

\end{itemize}

\begin{table*}[!htbp]\normalsize
    \centering
    \vspace{-2mm}
    \caption{Performance comparison of different methods on {the} GTEA dataset based on F-score.}
    \label{table:GTEA}
    \begin{small}
    \begin{tabular}{Il Ic Lc Lc Ic Ic}
    \whline
    \multirow{2}{*}{\bf Method} & \multicolumn{3}{cI}{\bf F-score} &
    \multicolumn{1}{cI}{\multirow{2}{*}{\bf Speed}}\\
    \cline{2-4}
   & coffee & tea & peanut & \multicolumn{1}{cI}{}\\
    \whline
    Single-pixel color~\cite{jones2002statistical} & 0.837 & 0.804 & 0.730 & $\sim$7fps\\
    \mhline
    Video stabilization~\cite{hayman2003statistical} & 0.376 & 0.305 & 0.310 & -\\
    \mhline
    Foreground modeling~\cite{sheikh2009background} & 0.275 & 0.239 & 0.255 & -\\
    \mhline
    Stabilization + CRF~\cite{fathi2011learning} & 0.713 & 0.812 & 0.727 & -\\
    \mhline
    Mixture of local and global~\cite{li2013pixel} & 0.933 & 0.943 & 0.883 & $\sim$10fps\\
    \mhline
    Deep-based approach~\cite{bambach2015lending} & 0.951 & 0.955 & \textbf{0.918} & $\sim$2fps\\
    \whline
    \textbf{Ours} & \textbf{0.952} & \textbf{0.959} & 0.911 & \textbf{$\sim$32fps}\\
    \whline
    \end{tabular}
    \end{small}
    \vspace{-2mm}
\end{table*}

\section{Experiments}\label{sec:experiments}

{In this section, we first introduce the datasets and the evaluation metrics in Section~\ref{sec:experiments_dataset}. We then evaluate the performance of the proposed method on different datasets in Sections~\ref{sec:experiments_result_GTEA}, \ref{sec:experiments_result_ADL} and \ref{sec:experiments_result_EgoHands},} followed by the ablation study in Section~\ref{sec:experiments_ablation} and the parameter analysis in Section~\ref{sec:experiments_parameter}. Finally, we analyze the efficiency of the proposed method in Section~\ref{sec:experiments_runtime}.

\subsection{Datasets and Evaluation Metrics}\label{sec:experiments_dataset}
{We evaluate the proposed method on {three datasets: the GTEA dataset~\cite{fathi2011learning}, the ADL} dataset~\cite{pirsiavash2012detecting} and the EgoHands dataset~\cite{bambach2015lending}.} These datasets and the evaluation metrics are summarized as follows:
\begin{itemize}
\item \textbf{GTEA dataset}: It is created primarily as an egocentric activity recognition dataset and contains 7 types of daily activities, each of which is performed by 4 different camera wearers. The videos are taken in the same environment under static illumination. {As only the wearer's hands are labeled with foreground hand masks in GTEA, social interactions with partners are not considered}. We evaluate the performance on the GTEA dataset following the settings introduced in~\cite{li2013pixel}, by classifying all videos into coffee, tea and peanut categories. F-score is used to quantify {the} classification performance.
\item \textbf{ADL dataset}: It contains 1 million frames of people performing everyday activities and is annotated with activities, object tracks, hand positions, and interaction events. It is challenging as it contains two types of actions: (1) long-duration actions, e.g., making tea that takes a few minutes, and (2) complex object interactions, e.g., opening the door of a fridge resulting in very different visual appearance. We evaluate the performance on the ADL dataset following the settings introduced in~\cite{li2013model}, and use F-score to quantify the performance.
\item {\textbf{EgoHands dataset}: it contains 48 Google Glass videos of complex, first-person interactions between two persons, i.e., the wearer and the partner, for 4,800 frames and more than 15,000 hands. The captured videos include realistic and challenging social situations where multiple hands appear. Four types of activities are involved: playing cards, playing chess, solving a jigsaw puzzle and playing Jenga. These activities are captured in three locations, and four camera wearers contribute to the dataset, {resulting in a total of $4 \times 3 \times 4 = 48$ videos}. In {this dataset,} the partner's hands appear in the vast majority of frames (95.2\% and 94.0\% for left and right), while the wearer's hands appear less often (53.3\% and 71.1\% for left and right). {This shows that} the wearer's hands are more frequently outside the visual field, and people tend to align attention with the dominant hand. We evaluate the performance on {this dataset following the settings introduced in~\cite{bambach2015lending}: intersection over union (IoU) between the estimated hand regions and the ground truth annotations}.}
\end{itemize}

{We further compare the runtime with {the} baseline methods on all datasets, by considering {the number of processed} frames per second (fps). As most methods do not report this information except for the method proposed in ~\cite{jones2002statistical}, we either directly {ran the available codes (for the method proposed in ~\cite{bambach2015lending}) or asked  the authors to help do it (for the methods proposed in ~\cite{li2013pixel,li2013model}). Failing these, we leave the runtime blank (for the methods proposed in ~\cite{hayman2003statistical,sheikh2009background,fathi2011learning}).}}

\subsection{Performance Comparison on {the} GTEA Dataset}\label{sec:experiments_result_GTEA}
We compare the proposed {method} with six methods on {the} GTEA dataset~\cite{fathi2011learning}: (1) the single-pixel color approach~\cite{jones2002statistical} that utilizes a random regressor, (2) the video stabilization approach~\cite{hayman2003statistical} based on background modeling by aligning a short sequence with affine transformation, (3) the foreground modeling~\cite{sheikh2009background} using feature trajectory-based projection which benefits from the KLT tracker~\cite{lucas1981iterative}, (4) the hybrid approach~\cite{fathi2011learning} with the combination of video stabilization and CRF, (5) the local and global appearance mixture model~\cite{li2013pixel} that highlights the effectiveness of sparse features and the importance of modeling global illumination, {and} (6) the deep-based approach combined with strong appearance models~\cite{bambach2015lending}. Partial results for these methods have been reported in~\cite{li2013pixel}. Table~\ref{table:GTEA} summarizes the experimental results for comparison.

\begin{figure*}[!htbp]
\centering
\subfigure{
\includegraphics[width=0.22\textwidth]{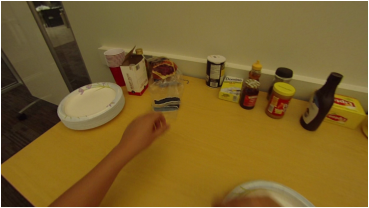}
}
\hspace{-3mm}
\subfigure{
\includegraphics[width=0.22\textwidth]{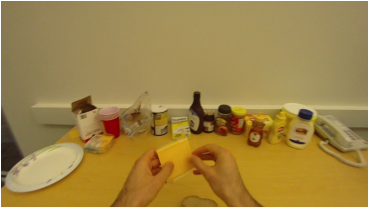}
}
\hspace{-3mm}
\subfigure{
\includegraphics[width=0.22\textwidth]{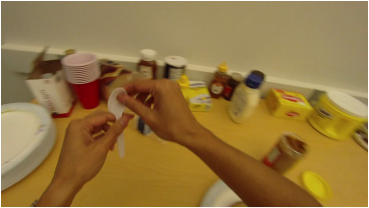}
}
\hspace{-3mm}
\subfigure{
\includegraphics[width=0.22\textwidth]{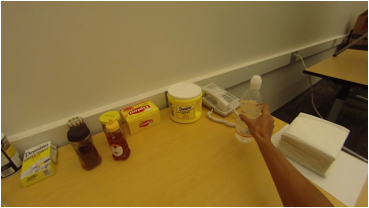}
}
\\
\vspace{-2mm}
\subfigure{
\includegraphics[width=0.22\textwidth]{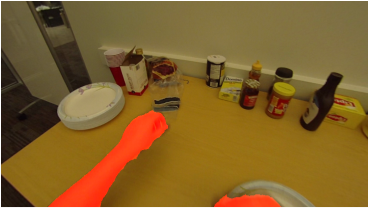}
}
\hspace{-3mm}
\subfigure{
\includegraphics[width=0.22\textwidth]{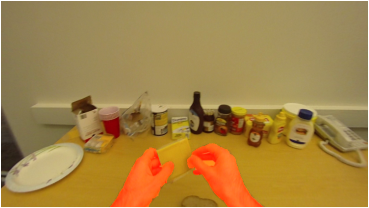}
}
\hspace{-3mm}
\subfigure{
\includegraphics[width=0.22\textwidth]{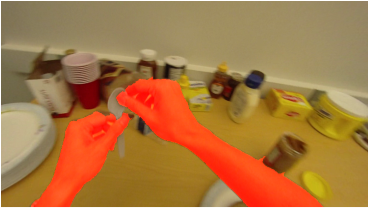}
}
\hspace{-3mm}
\subfigure{
\includegraphics[width=0.22\textwidth]{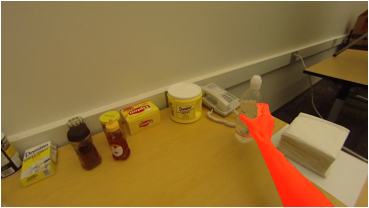}
}
\caption{{Qualitative results on the GTEA dataset. Top: original frames. Bottom:} detected hands (red).}
\label{fig:result_GTEA}
\end{figure*}

The single-pixel color approach~\cite{jones2002statistical} achieves {surprisingly good performance even though it was proposed nearly 15 years} ago. The video stabilization~\cite{hayman2003statistical} and the foreground modeling~\cite{sheikh2009background} approaches perform poorly as they were designed for hand-held videos instead of egocentric videos. {This demonstrates that there are indeed significant differences between the two types of videos,} and simply applying the traditional methods on egocentric videos would result in unsatisfactory results. {The fourth and the fifth methods~\cite{fathi2011learning,li2013pixel}, which are proposed recently for egocentric hand detection, perform generally well.
The deep-based approach~\cite{bambach2015lending} performs the best among all existing methods. Our method perform slightly better on coffee and tea but slightly worse on peanut. This shows that our method is comparable to the deep-based approach. However, in terms of efficient, our method performs much faster than the deep-based approach. In fact, ours is the fastest among all methods. {This experiment demonstrates that the proposed method performs well in terms of accuracy and efficiency on the GTEA dataset~\cite{fathi2011learning}, which contains videos of mostly static illumination with small camera motion and no social interactions.} The detection errors from the proposed method are mainly due to over-segmentation, especially for the fingers as they are easily confused with other objects and the background. Fig.~\ref{fig:result_GTEA} shows some qualitative results on the GTEA dataset~\cite{fathi2011learning}.}

\begin{figure*}[!htbp]
\centering
\subfigure{
\includegraphics[width=0.22\textwidth]{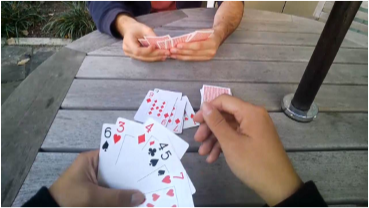}
}
\hspace{-3mm}
\subfigure{
\includegraphics[width=0.22\textwidth]{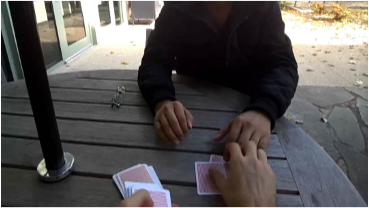}
}
\hspace{-3mm}
\subfigure{
\includegraphics[width=0.22\textwidth]{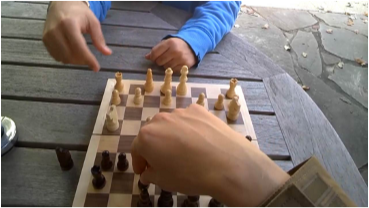}
}
\hspace{-3mm}
\subfigure{
\includegraphics[width=0.22\textwidth]{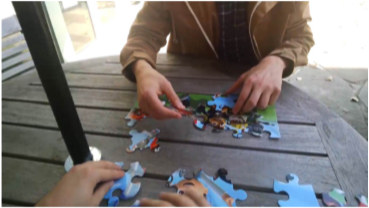}
}
\subfigure{
\includegraphics[width=0.22\textwidth]{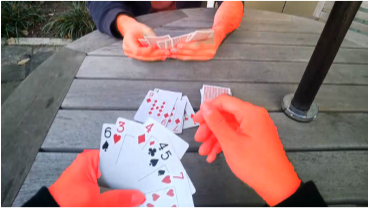}
}
\hspace{-3mm}
\subfigure{
\includegraphics[width=0.22\textwidth]{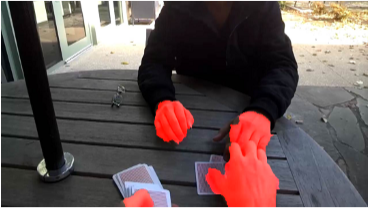}
}
\hspace{-3mm}
\subfigure{
\includegraphics[width=0.22\textwidth]{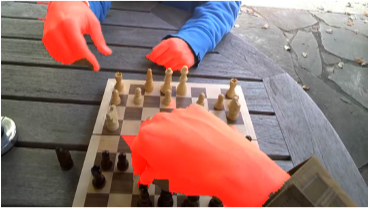}
}
\hspace{-3mm}
\subfigure{
\includegraphics[width=0.22\textwidth]{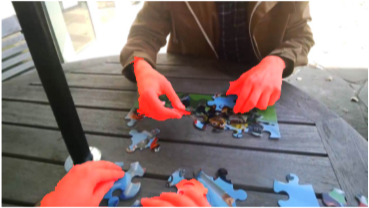}
}
\caption{Qualitative results on the EgoHands dataset. Top: original frames; bottom: hand detection results (red).}
\label{fig:result_EgoHands}
\end{figure*}

\subsection{Performance Comparison on the ADL Dataset}\label{sec:experiments_result_ADL}
We compare the proposed {method} with two baseline methods on the ADL dataset~\cite{pirsiavash2012detecting}, i.e., the local and global appearance mixture model~\cite{li2013pixel} and the model recommendation approach with virtual probes~\cite{li2013model} {and using different recommendation strategies}. Results for {these} baselines have been reported in~\cite{li2013model}, and Table~\ref{table:ADL} summarizes the experimental results for comparison.

\begin{table}[h]\normalsize
    \centering
    \caption{Performance comparison of different methods on the ADL dataset based on F-score.}
    \label{table:ADL}
    \begin{small}
    \begin{tabular}{Il Ic Ic Ic}
    \whline
    {\bf Method} & {\bf F-score} & {\bf Speed}\\
    \whline
    Mixture of local and global~\cite{li2013pixel} & 0.346 & $\sim$10fps\\
    \mhline
    Model recommendation (NMF)~\cite{li2013model} & 0.322 & $\sim$1fps\\
    \mhline
    Model recommendation (SC)~\cite{li2013model} & 0.252 & $\sim$1fps\\
    \mhline
    Model recommendation (KNN)~\cite{li2013model} & \textbf{0.384} & $\sim$1fps\\
    \mhline
    Model recommendation (RF)~\cite{li2013model} & 0.357 & $\sim$1fps\\
    \whline
    \textbf{Ours} & 0.375 & \textbf{$\sim$32fps}\\
    \whline
    \end{tabular}
    \end{small}
\end{table}

We can see that the proposed method outperforms the mixture model~\cite{li2013pixel}, which emphasizes the effectiveness of sparse features and the importance of modeling global illumination. The proposed method also outperforms the model recommendation approach~\cite{li2013model} when NMF, SC or RF is used as the recommendation strategy. On the other hand, its performance is comparable to it when KNN is used, which applies a non-linear model to capture complex feature mapping. However, the proposed method is again the most efficient method. {This experiment demonstrates that the proposed method achieves the favorable performance with higher efficiency compared with {the} baselines on the ADL dataset, which contains videos of real-life scenarios {with complex object interactions} in daily activities.}

\subsection{Performance Comparison on the EgoHands Dataset}\label{sec:experiments_result_EgoHands}
We compare the proposed approach with two baseline methods on the EgoHands dataset~\cite{bambach2015lending}, i.e., the local and global appearance mixture model~\cite{li2013pixel}, and the recently proposed deep-based  approach~\cite{bambach2015lending} combining with strong appearance models. In the experiments, we neither conduct the activity recognition, nor carry out the classification of ``the left hand'' or ``the right hand'', and only compare the results of hand segmentation. Table~\ref{table:EgoHands} shows the experimental results with the same setting used in~\cite{bambach2015lending}, and Fig.~\ref{fig:result_EgoHands} shows some qualitative results.

\begin{table}\normalsize
    \centering
    \caption{Comparative results on EgoHands dataset. IoU score against baseline methods}
    \label{table:EgoHands}
    \begin{small}
    \begin{tabular}{Il Ic Ic Ic}
    \whline
    {\bf Method} & {\bf IoU score} & {\bf Speed} \\
    \whline
    Mixture of local and global~\cite{li2013pixel} & 0.478  & $\sim$6fps\\
    \mhline
    Deep-based approach~\cite{bambach2015lending} & \textbf{0.556}  & $\sim$1fps\\
    \mhline
    \textbf{Ours} & 0.527  & \textbf{$\sim$18fps}\\
    \whline
    \end{tabular}
    \end{small}
\end{table}

This work achieves a better result compared with~\cite{li2013pixel} which is based on hand-crafted cues like ours, while performs a little worse than the deep-based approach~\cite{bambach2015lending} with much higher efficiency for the following reasons:

- ~\cite{bambach2015lending} starts from the pre-trained CaffeNet, indicating it benefits from the additional data thus unfair for comparison.

- Different from other works~\cite{li2013pixel,li2013model}, the EgoHands dataset defines the ``hand'' to stop at the wrist instead of the sleeve. Similar to the failures of~\cite{li2013pixel}, the proposed method cannot effectively distinguish hands and arms, as they share similar motions and appearances. However, \cite{bambach2015lending} succeeds in these cases by learning from sufficient labeled samples.

- The proposed method would generate inaccurate seeds when the moving bystanders or objects appear in the background, as we assume that the motions are mainly produced by the mounted camera and the hands, which is not satisfied in the above scenario.

\begin{table*}[!htbp]\normalsize
    \centering
    \caption{Ablation study on {the} GTEA, ADL and EgoHands datasets.}
    \label{table:ablation}
    \begin{small}
    \begin{tabular}{Il Ic Lc Lc Ic Ic Ic}
    \whline
    \multirow{2}{*}{\diagbox{\bf Method}{\bf Dataset}} & \multicolumn{3}{cI}{\bf GTEA} & \multicolumn{1}{cI}{\multirow{2}{*}{\bf ADL}} & \multicolumn{1}{cI}{\multirow{2}{*}{\bf EgoHands}}\\
    \cline{2-4}
   & coffee & tea & peanut & \multicolumn{1}{cI}{} & \multicolumn{1}{cI}{}\\
    \whline
    No ``contrast'' & 0.762 & 0.768 & 0.728 & 0.305 & 0.467\\
    \mhline
    No ``location'' & 0.914 & 0.921 & 0.872 & 0.334 & 0.486\\
    \mhline
    No ``position consistency'' & 0.876 & 0.883 & 0.836 & 0.347 & 0.492\\
    \mhline
    No ``appearance continuity'' & 0.791 & 0.796 & 0.755 & 0.316 & 0.448\\
    \whline
    \textbf{Ours} & \textbf{0.952} & \textbf{0.959} & \textbf{0.911} & \textbf{0.375} & \textbf{0.527}\\
    \whline
    \end{tabular}
    \end{small}
\end{table*}

\subsection{Ablation Study}\label{sec:experiments_ablation}

To evaluate the impacts of {the proposed egocentric cues (Section~\ref{sec:method_growth}), we have conducted an experiment to compare different versions of the proposed method, with each version having one cue (i.e., one term in Eq.~\ref{equ:fusion}) removed. Table~\ref{table:ablation} shows the results on the three datasets. From these results, we can draw the following conclusions:}
\begin{itemize}
\item The appearance {constraints, i.e., contrast and appearance continuity,} are observed to have larger impacts compared with the spatial {constraints, i.e., location and position consistency, on the three datasets. This indicates that the skin color plays an important role on} egocentric hand detection.

\item For {the GTEA dataset, the appearance constraints are more critical as the illumination in the videos tends to be static, meaning that the contrast and appearance continuity cues are more reliable here.}

\item For the ADL dataset, the impacts of the appearance constraints decrease, due to the wider range of indoor imaging conditions, while the impacts of the spatial constraints increase, due to the long-duration actions and complex object interactions (which require more accurate hand positioning and tracking).

\item {For the EgoHands dataset, ``appearance continuity'' stands out from all cues {with more models constructed} compared with the GTEA dataset, as multiple hands are involved in social interactions. ``Location'' and ``position consistency'' also affect the performance because of the strict definition of hands.}

\end{itemize}

\subsection{Parameter Analysis}\label{sec:experiments_parameter}

In Algorithm~\ref{alg:at_growth}, {the} hand regions are gradually located from the seed regions with two parameters. {While $\alpha$ defines the termination condition of region growing, $\beta$ defines the threshold for noise component removal. In general:}
\begin{itemize}
\item A large $\alpha$ {value increases the rejection rate. Thus, fewer superpixels are retained, resulting in the incompleteness of the hand regions, e.g., the missing fingers. On} the other hand, a small $\alpha$ may {produce some} redundant regions that are hand-irrelevant, e.g., the grasped bread.

\item A large $\beta$ {value increases the risk of falsely removing components that belong to the hands, e.g., the partially occluded palm, while a small $\beta$ may not be effectively enough to exclude noisy} components, e.g., the chess.
\end{itemize}

To {study} how the results are affected by $\alpha$ and $\beta$, we compare the performances {of having different combinations of $\alpha$ and $\beta$ on the ADL dataset}. Table~\ref{table:Parameter} {shows} the comparison results, which agrees with the above {analysis}. {In addition, these results show that within an appropriate range of $\alpha$ and $\beta$, there is only a small change in performance, indicating that the proposed method is robust} to parameter variations.

\begin{table}[h]\normalsize
    \centering
    \vspace{-2mm}
    \caption{Parameter analysis by adopting various combinations of $\alpha$ and $\beta$ on {the} ADL dataset.}
    \label{table:Parameter}
    \begin{small}
    \begin{tabular}{Ic Ic Lc Lc Lc Lc Ic}
    \whline
    \diagbox{\bf $\alpha$}{\bf $\beta$} & 300 & 350 & 400 & 450 & 500\\
    \whline
    0.5 &  0.351  & 0.359  & 0.363  & 0.361  & 0.358 \\
    \mhline
    0.55 & 0.361  & 0.367  & 0.370  & 0.368  & 0.364 \\
    \mhline
    0.6 &  0.367  & 0.372  & \textbf{0.375}  & 0.373  & 0.368 \\
    \mhline
    0.65 & 0.363  & 0.368  & 0.371  & 0.368  & 0.362 \\
    \mhline
    0.7 &  0.357  & 0.361  & 0.364  & 0.360  & 0.352 \\
    \whline
    \end{tabular}
    \end{small}
    \vspace{-2mm}
\end{table}

\subsection{Runtime Analysis}\label{sec:experiments_runtime}
{We have implemented the proposed {method} on a PC with an i7 2.6GHz CPU and 16GB RAM. The average rate is $\sim$32fps, indicating that it can run in} real-time (640 $\times$ 360). {The region growing step takes more time compared with the seed generation step ($\sim$21ms vs. $\sim$10ms). Without the early rejection step discussed in Section~\ref{sec:method_discussion} to avoid unnecessarily computing all cues, an additional $\sim$23ms would be needed. The early rejection step only causes a 1.3\% drop in performance on the} ADL dataset. Compared with {the existing methods, a significant speedup is observed, e.g., $\sim$1s using the model recommendation approach~\cite{li2013model} (640 $\times$ 360, on a 2.6GHz CPU), $\sim$100ms using the local and global appearance mixture model~\cite{li2013pixel} (640 $\times$ 360, on a 2.6GHz CPU), $\sim$2s using DP-DPM, and 9s using R-CNN as reported in ~\cite{deng2016joint} (500 $\times$ 400, on a 2.9GHz CPU and Titan X GPU), $\sim$40ms using region growing as reported in ~\cite{kumar2015fly} (320 $\times$ 180, on a 2.80GHz CPU), $\sim$29ms using YOLO growing as reported in ~\cite{rangesh2016driver} (640 $\times$ 480, on a 3.5GHz 6-core CPU and Titan X GPU).}

{We {have further conducted an experiment to study the efficiency of the proposed method on two older PCs with limited computational capabilities and memory space, to simulate the computation environment of wearable devices. The following experimental setups are tested:
\begin{enumerate}
\item A PC with Pentium IV 2.53 GHZ CPU and 512M RAM, video resolution at $320 \times 240$: We achieve a framerate of $\sim$13fps.
\item A PC with i3 3.3 GHZ CPU and 2G RAM, video resolution at $320 \times 240$: We achieve a framerate of $\sim$38fps.
\item A PC with i3 3.3 GHZ CPU and 2G RAM, video resolution at $640 \times 360$: We achieve a framerate of $\sim$15fps.
\end{enumerate}
These results indicate the high efficiency of the proposed method, even when compared with relevant works in virtual reality~\cite{gencc2015handvr} and augmented reality~\cite{liang2015ar}.}}

\section{Conclusion}\label{sec:conclusion}
In this paper, we have proposed a seed region generation algorithm and novel egocentric cues for egocentric hand detection, by growing the detected seed regions into hand regions iteratively. We have discussed some implementation issues in applying the proposed method in complex environments. Experimental evaluations on public datasets demonstrate that the proposed method performs comparably against the state-of-the-art methods, while being significantly more efficient. Ablation study and parameter analysis have been conducted to study the impacts of the proposed cues and the important parameters. As a future work, we are currently applying the proposed method on some related tasks like gesture recognition and hand-based action recognition.

\ifCLASSOPTIONcaptionsoff
  \newpage
\fi

\bibliographystyle{IEEEtran}
\bibliography{IEEEabrv,egbib}

\begin{thebibliography}{10}
\providecommand{\url}[1]{#1}
\csname url@samestyle\endcsname
\providecommand{\newblock}{\relax}
\providecommand{\bibinfo}[2]{#2}
\providecommand{\BIBentrySTDinterwordspacing}{\spaceskip=0pt\relax}
\providecommand{\BIBentryALTinterwordstretchfactor}{4}
\providecommand{\BIBentryALTinterwordspacing}{\spaceskip=\fontdimen2\font plus
\BIBentryALTinterwordstretchfactor\fontdimen3\font minus
  \fontdimen4\font\relax}
\providecommand{\BIBforeignlanguage}[2]{{%
\expandafter\ifx\csname l@#1\endcsname\relax
\typeout{** WARNING: IEEEtran.bst: No hyphenation pattern has been}%
\typeout{** loaded for the language `#1'. Using the pattern for}%
\typeout{** the default language instead.}%
\else
\language=\csname l@#1\endcsname
\fi
#2}}
\providecommand{\BIBdecl}{\relax}
\BIBdecl

\bibitem{li2013learning}
Y.~Li, A.~Fathi, and J.~Rehg, ``Learning to predict gaze in egocentric video,''
  in \emph{ICCV}, 2013, pp. 3216--3223.

\bibitem{kumano2016}
S.~Kumano, K.~Otsuka, R.~Ishii, and J.~Yamato, ``Collective first-person vision
  for automatic gaze analysis in multiparty cnversations,'' \emph{IEEE Trans.
  on Multimedia}, vol.~19, no.~1, pp. 107--122, 2017.

\bibitem{ghosh2012discovering}
J.~Ghosh, Y.~J. Lee, and K.~Grauman, ``Discovering important people and objects
  for egocentric video summarization,'' in \emph{CVPR}, 2012, pp. 1346--1353.

\bibitem{lu2013story}
Z.~Lu and K.~Grauman, ``Story-driven summarization for egocentric video,'' in
  \emph{CVPR}, 2013, pp. 2714--2721.

\bibitem{dominguez2006robust}
S.~M. Dominguez, T.~Keaton, and A.~H. Sayed, ``A robust finger tracking method
  for multimodal wearable computer interfacing,'' \emph{IEEE Trans. on
  Multimedia}, vol.~8, no.~5, pp. 956--972, 2006.

\bibitem{xiong2014detecting}
B.~Xiong and K.~Grauman, ``Detecting snap points in egocentric video with a web
  photo prior,'' in \emph{ECCV}, 2014, pp. 282--298.

\bibitem{Wang_2016_CVPR}
J.~Wang, Y.~Cheng, and R.~Schmidt~Feris, ``Walk and learn: facial attribute
  representation learning from egocentric video and contextual data,'' in
  \emph{CVPR}, 2016, pp. 2295--2304.

\bibitem{Hoshen_2016_CVPR}
Y.~Hoshen and S.~Peleg, ``An egocentric look at video photographer identity,''
  in \emph{CVPR}, 2016, pp. 4284--4292.

\bibitem{fathi2011learning}
A.~Fathi, X.~Ren, and J.~M. Rehg, ``Learning to recognize objects in egocentric
  activities,'' in \emph{CVPR}, 2011, pp. 3281--3288.

\bibitem{li2013model}
C.~Li and K.~Kitani, ``Model recommendation with virtual probes for egocentric
  hand detection,'' in \emph{ICCV}, 2013, pp. 2624--2631.

\bibitem{bambach2015lending}
S.~Bambach, S.~Lee, D.~J. Crandall, and C.~Yu, ``Lending a hand: detecting
  hands and recognizing activities in complex egocentric interactions,'' in
  \emph{ICCV}, 2015, pp. 1949--1957.

\bibitem{li2013pixel}
C.~Li and K.~Kitani, ``Pixel-level hand detection in ego-centric videos,'' in
  \emph{CVPR}, 2013, pp. 3570--3577.

\bibitem{ren2010figure}
X.~Ren and C.~Gu, ``Figure-ground segmentation improves handled object
  recognition in egocentric video,'' in \emph{CVPR}, 2010, pp. 3137--3144.

\bibitem{lee2014hand}
S.~Lee, S.~Bambach, D.~Crandall, J.~Franchak, and C.~Yu, ``This hand is my
  hand: a probabilistic approach to hand disambiguation in egocentric video,''
  in \emph{CVPR Workshops}, 2014, pp. 543--550.

\bibitem{betancourt2015evolution}
A.~Betancourt, P.~Morerio, C.~S. Regazzoni, and M.~Rauterberg, ``The evolution
  of first person vision methods: A survey,'' \emph{TCSVT}, vol.~25, no.~5, pp.
  744--760, 2015.

\bibitem{bambach2015survey}
S.~Bambach, ``A survey on recent advances of computer vision algorithms for
  egocentric video,'' \emph{arXiv:1501.02825}, 2015.

\bibitem{del2017summarization}
A.~G. del Molino, C.~Tan, J.-H. Lim, and A.-H. Tan, ``Summarization of
  egocentric videos: A comprehensive survey,'' \emph{IEEE Trans. on
  Human-Machine Systems}, vol.~47, no.~1, pp. 65--76, 2017.

\bibitem{baraldi2014gesture}
L.~Baraldi, F.~Paci, G.~Serra, L.~Benini, and R.~Cucchiara, ``Gesture
  recognition in ego-centric videos using dense trajectories and hand
  segmentation,'' in \emph{CVPR Workshops}, 2014, pp. 688--693.

\bibitem{rogez2015first}
G.~Rogez, J.~S. Supancic, and D.~Ramanan, ``First-person pose recognition using
  egocentric workspaces,'' in \emph{CVPR}, 2015, pp. 4325--4333.

\bibitem{li2015delving}
Y.~Li, Z.~Ye, and J.~M. Rehg, ``Delving into egocentric actions,'' in
  \emph{CVPR}, 2015, pp. 287--295.

\bibitem{achanta2012slic}
R.~Achanta, A.~Shaji, K.~Smith, A.~Lucchi, P.~Fua, and S.~Susstrunk, ``{SLIC}
  superpixels compared to state-of-the-art superpixel methods,'' \emph{TPAMI},
  vol.~34, no.~11, pp. 2274--2282, 2012.

\bibitem{kolsch2004fast}
M.~Kolsch and M.~Turk, ``Fast {2D} hand tracking with flocks of features and
  multi-cue integration,'' in \emph{CVPR Workshops}, 2004, pp. 158--158.

\bibitem{pirsiavash2012detecting}
H.~Pirsiavash and D.~Ramanan, ``Detecting activities of daily living in
  first-person camera views,'' in \emph{CVPR}, 2012, pp. 2847--2854.

\bibitem{fathi2012learning}
A.~Fathi, Y.~Li, and J.~M. Rehg, ``Learning to recognize daily actions using
  gaze,'' in \emph{ECCV}, 2012, pp. 314--327.

\bibitem{rublee2011orb}
E.~Rublee, V.~Rabaud, K.~Konolige, and G.~Bradski, ``{ORB}: an efficient
  alternative to {SIFT} or {SURF},'' in \emph{ICCV}, 2011, pp. 2564--2571.

\bibitem{lowe2004distinctive}
D.~G. Lowe, ``Distinctive image features from scale-invariant keypoints,''
  \emph{IJCV}, vol.~60, no.~2, pp. 91--110, 2004.

\bibitem{bay2006surf}
H.~Bay, T.~Tuytelaars, and L.~Van~Gool, ``{SURF}: speeded up robust features,''
  in \emph{ECCV}, 2006, pp. 404--417.

\bibitem{fischler1981random}
M.~A. Fischler and R.~C. Bolles, ``Random sample consensus: a paradigm for
  model fitting with applications to image analysis and automated
  cartography,'' \emph{Communications of the ACM}, vol.~24, no.~6, pp.
  381--395, 1981.

\bibitem{uijlings2013selective}
J.~R. Uijlings, K.~E. van~de Sande, T.~Gevers, and A.~W. Smeulders, ``Selective
  search for object recognition,'' \emph{IJCV}, vol. 104, no.~2, pp. 154--171,
  2013.

\bibitem{cevikalp2013face}
H.~Cevikalp, B.~Triggs, and V.~Franc, ``Face and landmark detection by using
  cascade of classifiers,'' in \emph{Automatic Face and Gesture Recognition},
  2013, pp. 1--7.

\bibitem{betancourt2014sequential}
A.~Betancourt, M.~Lopez, C.~Regazzoni, and M.~Rauterberg, ``A sequential
  classifier for hand detection in the framework of egocentric vision,'' in
  \emph{CVPR Workshops}, 2014, pp. 586--591.

\bibitem{jones2002statistical}
M.~J. Jones and J.~M. Rehg, ``Statistical color models with application to skin
  detection,'' \emph{IJCV}, vol.~46, no.~1, pp. 81--96, 2002.

\bibitem{hayman2003statistical}
E.~Hayman and J.-O. Eklundh, ``Statistical background subtraction for a mobile
  observer,'' in \emph{ICCV}, 2003, pp. 67--74.

\bibitem{sheikh2009background}
Y.~Sheikh, O.~Javed, and T.~Kanade, ``Background subtraction for freely moving
  cameras,'' in \emph{ICCV}, 2009, pp. 1219--1225.

\bibitem{lucas1981iterative}
B.~D. Lucas, T.~Kanade \emph{et~al.}, ``An iterative image registration
  technique with an application to stereo vision.'' in \emph{IJCAI}, vol.~81,
  no.~1, 1981, pp. 674--679.

\bibitem{deng2016joint}
X.~Deng, Y.~Yuan, Y.~Zhang, P.~Tan, L.~Chang, S.~Yang, and H.~Wang, ``Joint
  hand detection and rotation estimation by using {CNN},''
  \emph{arXiv:1612.02742}, 2016.

\bibitem{kumar2015fly}
J.~Kumar, Q.~Li, S.~Kyal, E.~A. Bernal, and R.~Bala, ``On-the-fly hand
  detection training with application in egocentric action recognition,'' in
  \emph{CVPR Workshops}, 2015, pp. 18--27.

\bibitem{rangesh2016driver}
A.~Rangesh, E.~Ohn-Bar, M.~M. Trivedi \emph{et~al.}, ``Driver hand localization
  and grasp analysis: a vision-based real-time approach,'' in \emph{ITSC},
  2016, pp. 2545--2550.

\bibitem{gencc2015handvr}
S.~Gen{\c{c}}, M.~Ba{\c{s}}tan, U.~G{\"u}d{\"u}kbay, V.~Atalay, and
  {\"O}.~Ulusoy, ``{HandVR}: a hand-gesture-based interface to a video
  retrieval system,'' \emph{Signal, Image and Video Processing}, vol.~9, no.~7,
  pp. 1717--1726, 2015.

\bibitem{liang2015ar}
H.~Liang, J.~Yuan, D.~Thalmann, and N.~M. Thalmann, ``{AR} in hand: Egocentric
  palm pose tracking and gesture recognition for augmented reality
  applications,'' in \emph{ACM Multimedia}, 2015, pp. 743--744.

\end{thebibliography}





\end{document}